\documentclass[10pt,twocolumn,letterpaper]{article}

\usepackage{multirow}
\usepackage{makecell}
\usepackage{tcolorbox}
\usepackage{amsthm}

\usepackage[pagenumbers]{cvpr} % To force page numbers, e.g. for an arXiv version

\definecolor{cvprblue}{rgb}{0.21,0.49,0.74}
\usepackage[pagebackref,breaklinks,colorlinks,allcolors=cvprblue]{hyperref}

\def\paperID{} % *** Enter the Paper ID here
\def\confName{CVPR}
\def\confYear{2026}

\title{Concise Geometric Description as a Bridge: \\ Unleashing the Potential of LLM for Plane Geometry Problem Solving}

\author{
 \textbf{Jingyun Wang\textsuperscript{1}},
 \textbf{Dian Li\footnotemark[3]},
 \textbf{Xiaohan Wang},
 \textbf{Gang Liu},
 \textbf{Jiahong Yan},
 \textbf{Guoliang Kang \textsuperscript{1}\footnotemark[3]}
\\
\\
 \textsuperscript{1}Beihang University
\\
   \texttt{\{wangjingyun0730, kgl.prml\}@gmail.com}\\
}

\begin{document}
\maketitle

\renewcommand{\thefootnote}{\fnsymbol{footnote}}
% \footnotetext[2]{Project leader.}
\footnotetext[3]{Corresponding authors.}

\begin{abstract}
Plane Geometry Problem Solving (PGPS) is a multimodal reasoning task that aims to solve a plane geometric problem based on a geometric diagram and problem textual descriptions.
Although Large Language Models (LLMs) possess strong reasoning skills, their direct application to PGPS is hindered by their inability to process visual diagrams. 
Existing works typically fine-tune Multimodal LLMs (MLLMs) end-to-end on large-scale PGPS data to enhance visual understanding and reasoning simultaneously.
However, such joint optimization may compromise base LLMs' inherent reasoning capability.
In this work, we observe that LLM itself is potentially a powerful PGPS solver when appropriately formulating visual information as textual descriptions.
We propose to train a MLLM Interpreter to generate geometric descriptions for the visual diagram, and an off-the-shelf LLM is utilized to perform reasoning.
Specifically, we choose Conditional Declaration Language (CDL) as the geometric description as its conciseness eases the MLLM Interpreter training. 
The MLLM Interpreter is fine-tuned via CoT (Chain-of-Thought)-augmented SFT followed by GRPO to generate CDL.
Instead of using a conventional solution-based reward that compares the reasoning result with the ground-truth answer, we design CDL matching rewards to facilitate more effective GRPO training, which provides more direct and denser guidance for CDL generation.
To support training, we construct a new dataset, Formalgeo7k-Rec-CoT, by manually reviewing Formalgeo7k v2 and incorporating CoT annotations.
Extensive experiments on Formalgeo7k-Rec-CoT, Unigeo, and MathVista show our method (finetuned on only 5.5k data) performs favorably against leading open-source and closed-source MLLMs.

\end{abstract}

\section{Introduction}
\label{sec:intro}

\begin{figure*}
  \centering
  \includegraphics[width=\linewidth]{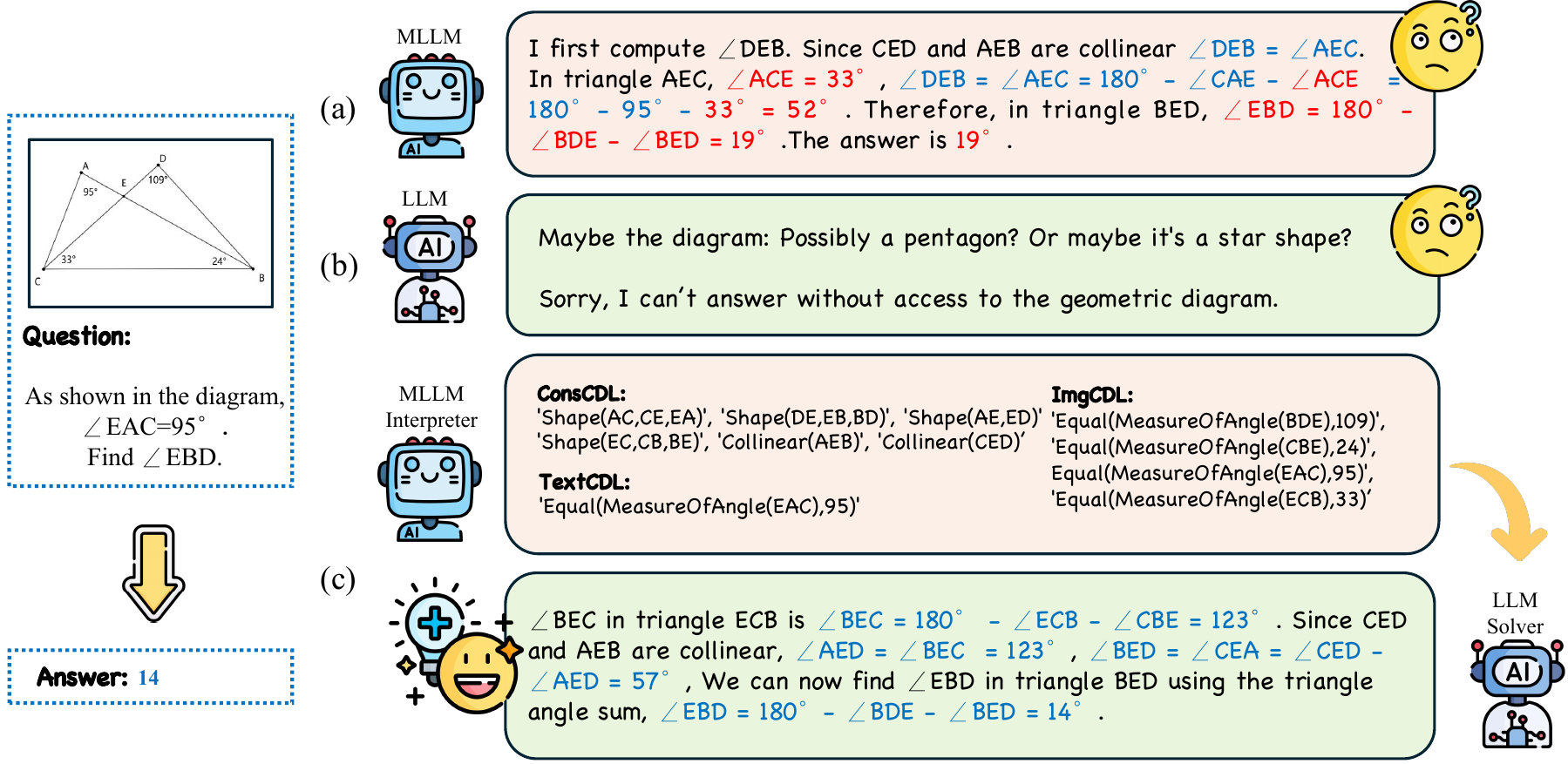}
  \caption{
  \textbf{(a) MLLMs.} Though MLLMs can directly perform PGPS, they still suffer from heavy visual perception errors or logical reasoning errors.
  \textbf{(b) LLMs. } LLMs are not capable of PGPS without access to geometric diagrams.  
  \textbf{(c) Ours.} We employ a MLLM interpreter to convert geometric diagrams into a concise CDL description upon which an LLM solver performs reasoning.}
\label{fig：intro}
\end{figure*}

Plane Geometry Problem Solving (PGPS) is a multimodal task that requires solving a problem based on a geometric diagram and its accompanying textual description. 
This task presents a significant challenge, as it demands both complex visual perception and rigorous logical reasoning capabilities.
Though Large Language Models (LLMs) possess powerful reasoning skills, their direct application to PGPS (Fig.~\ref{fig：intro}(b)) is severely constrained without access to the visual modality. 

\begin{table}
\setlength{\tabcolsep}{4pt}
\renewcommand{\arraystretch}{1}
  \caption{\textbf{LLMs are potentially powerful PGPS solvers.} 
  ``Visual'' refers to MLLMs reasoning with visual inputs, while ``Caption'' and ``GT CDL'' refer to MLLMs and LLM reasoning with captions generated by Gemini-2.5 Pro or ground-truth CDL annotations, but without access to visual modality.
  Results demonstrate that, converting visual information into appropriate textual descriptions, LLM itself is a potentially powerful PGPS solver.}

  \label{tab:intro_llm}
  \centering
  \begin{tabular}{l|c|ccc}
    \toprule[1pt]
    Models & Type & Visual & Caption & GT CDL \\
    \midrule
    Claude-Opus-4.1 &\multirow{2}{*}{\makecell[c]{MLLM}}  & $69.1$ & $76.8$ & $84.2$\\
    Gemini2.5-Pro &  & $81.8$ & $81.1$ & $84.3$\\
    \midrule
    Qwen3 30B & LLM&- & \textbf{82.1} & \textbf{88.4}\\
    \bottomrule[1pt]
  \end{tabular}
  \vspace{-2mm}
\end{table}

Conventional PGPS methods~\cite{Diagram-Formalizer,Unigeo,PGPSNet,autogps,G-LLaVA,GeoGen,FGeo-Parser} turn to fine-tuning Multi-modal Large Language Models~\cite{Qwen2.5-vl,GLM-V,Llava} (MLLMs) on large-scale datasets to simultaneously enhance their visual understanding and reasoning capabilities.
Despite the progress achieved, such a joint optimization paradigm remains unsatisfactory:
On one hand, in Figure~\ref{fig：intro}(a), MLLMs still suffer from significant visual perception errors (\emph{e.g.}, misidentifying $\angle{ACE}$ as $33^\circ$ instead of $\angle{ECB}$).
On the other hand, end-to-end training may inevitably compromise the inherent reasoning skills of the base LLMs.
Moreover, we observe that (Table~\ref{tab:intro_llm}) %demonstrate a key finding: 
\textbf{LLMs are potentially powerful PGPS solvers when formulating visual information as textual descriptions}.
The preceding analysis motivates us to explore a new paradigm:
converting geometric inputs into textual descriptions to unleash the reasoning power of LLMs for PGPS.

In this paper, we propose a novel paradigm for PGPS.
As illustrated in Figure~\ref{fig：intro}(c), we train a MLLM Interpreter to convert geometric inputs into concise geometric descriptions, \emph{i.e.,} Conditional Declaration Language (CDL)~\cite{formalgeo} (as shown in pink box of Fig.~\ref{fig：intro}(c)). % for the LLM solver's further reasoning.
The MLLM Interpreter is initialized with an open-source MLLM and fine-tuned via a two-stage pipeline: a CoT-augmented Supervised Fine-Tuning (SFT) stage followed by a Group Relative Policy Optimization~\cite{GRPO} (GRPO) stage.
Instead of using a conventional solution-based reward that compares the reasoning result with the ground-truth answer, we design CDL matching rewards to facilitate more effective GRPO training, which provides more direct and denser guidance for CDL generation.
Specifically, we greedily match generated CDL and the ground-truth CDL piece-by-piece and design parsing rules to enable accurate matching. The CDL rewards consist of the recall and precision of the matching results.
Finally, with the generated geometric descriptions as the bridge, an off-the-shelf LLM Solver is utilized to perform reasoning.
Our practice reveals two technical insights
1) Concise geometric descriptions ease the MLLM Interpreter's learning by narrowing the search space.
2) Compared to solution-based rewards, our CDL matching rewards design is more effective in GRPO as it provides more direct and denser guidance for CDL generation. 

To support the training, we construct a new dataset, Formalgeo7k-Rec-CoT.
We conduct a rigorous manual review on Formalgeo7k v2~\cite{FGeo-Parser} to fix annotation errors and parse CDL annotations to generate Chain-of-Thoughts (CoTs). %, which formalize the reasoning process behind CDL generation.
%The resulting MLLM interpreter demonstrates superior CDL generation performance, surpassing previous methods.
We conduct extensive experiments on various PGPS benchmarks, including Formalgeo-Rec-CoT, Unigeo~\cite{Unigeo}, and the geometric set of MathVista~\cite{mathvista}.
Our method is compared with various leading open-source and closed-source MLLMs. 
The results demonstrate that our method consistently outperforms all open-source MLLMs and obtains comparable performance to leading closed-source MLLMs.

In a nutshell, our contributions are summarized as:

\begin{itemize}[leftmargin=2em]
    \item We observe LLM itself is a powerful PGPS solver when provided geometric information appropriately. Thus, we propose a new paradigm that trains a MLLM Interpreter to generate geometric descriptions and utilizes an off-the-shelf LLM for reasoning.
    \item We train the MLLM Interpreter with CoT-augmented SFT and GRPO with specifically designed CDL matching rewards. Our practice reveals two technical insights 
    1) Concise geometric descriptions ease the MLLM Interpreter's learning by narrowing the search space.
    2) CDL matching rewards are more effective in GRPO as they provide more direct and denser guidance for CDL generation than solution-based rewards. 
    \item We propose Formalgeo7k-Rec-CoT by conducting a rigorous manual review on Formalgeo7k v2 to fix annotation errors and parsing annotations to generate CoTs to support CDL generation training. 
    \item With only $5.5$k training data, our method performs favorably against leading open-source and closed-source MLLMs on PGPS benchmarks.
\end{itemize}

\section{Related Work}
\label{sec:related_work}

\noindent \textbf{Multi-modal Large Language Model (MLLM). }
Building upon breakthroughs in Large Language Models (LLMs)~\cite{GPT-3,MoE,deepseek-r1,llama,llama2,chatgpt, ERM,Qwen3,Yi} and Vision Foundation Models (VFM)~\cite{CLIP,SAM,blip,dino}, Multi-modal Large Language Models (MLLMs)~\cite{Qwen2.5-vl,Gemini2.5,Gpt-4o,GLM-V,Llava,gpt5} emerge as a powerful paradigm, which integrates visual perception with the reasoning capabilities of LLMs.
Through massive-scale pre-training data and advanced training strategies~\cite{PPO,DPO,GRPO,DAPO}, MLLMs make great progress in a wide range of multi-modal tasks, including visual question answering and image captioning.
However, their performance remains unsatisfactory in domains requiring rigorous reasoning (\emph{e.g.}, Plane Geometry Problem Solving).
A key limitation is that the inherent reasoning ability of their base LLMs might be compromised as MLLMs jointly optimize their visual perception and linguistic reasoning capabilities.

\noindent \textbf{Plane Geometry Problem Solving (PGPS). }
Most methods for Plane Geometry Problem Solving (PGPS)~\cite{G-LLaVA,Geosense,geouni,mavis,eagle,PGPSNet,GeoQA,autogeo,trustgeogen,SlowPerception} fine-tune Multi-modal Large Language Models (MLLMs) in an end-to-end manner on large-scale training data.
For example, PGPSNet~\cite{PGPSNet} constructs PGPS9K and G-LLaVA constructs Geo170K for training.
However, such end-to-end fine-tuning inevitably compromises the inherent reasoning capabilities of the base Large Language Models (LLMs).
Another line of work, including GF Reasoner~\cite{GFReasoner} and LVDA~\cite{LVDA}, explores converting diagrams into textual descriptions for geometric solvers.
Despite the progress, these methods face significant limitations: 
From one aspect, their reliance on general textual descriptions introduces substantial redundancy in the search space, which harms both description generation and solving performance. 
From the other aspect, the unstructured nature of general textual descriptions makes matching rewards in GRPO infeasible, forcing reliance on an extra solver that provides an extremely sparse reward signal. 
Beyond previous works, our method transforms geometric inputs into concise geometric descriptions, effectively addressing these limitations and fully leveraging the reasoning power of LLMs for PGPS.

\section{Method}
\label{sec:method}

\begin{figure*}
  \centering
  \includegraphics[width=\linewidth]{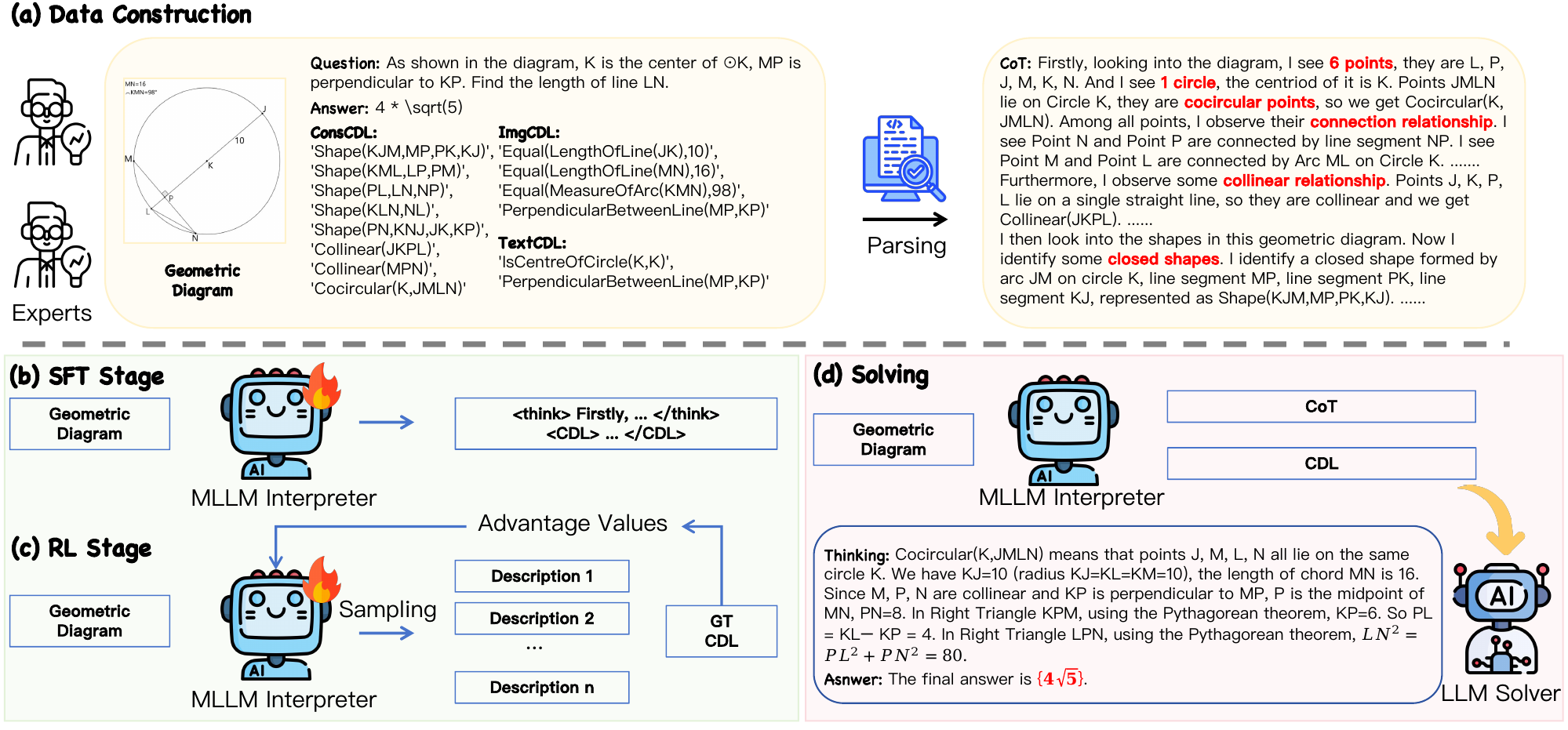}
  \caption{
  \textbf{Method Overview.}
  \textbf{(a) Data Construction.} We construct \textbf{Formalgeo7k-Rec-CoT} by manual reviewing Formalgeo7k v2 and incorporating the Chain-of-Thought (CoT).
  We design a two-stage pipeline to train MLLM Interpreter, including a \textbf{(b) CoT-Augmented SFT Stage} and a \textbf{(c) GRPO Stage with CDL Matching Rewards}.
  Based on the generated CDL, an \textbf{(d) LLM Solver} directly performs reasoning and derives final solutions.}
\label{fig：method}
\end{figure*}

\noindent \textbf{Overview}
In this work, we propose a new paradigm that firstly trains a MLLM Interpreter to generate geometric descriptions for the visual diagram, and then employs an off-the-shelf LLM to perform reasoning. The overall framework is illustrated in Fig.~\ref{fig：method}.
The MLLM Interpreter (Sec.~\ref{sec: MLLM Interpreter}) is trained with a two-stage pipeline to convert geometric diagrams into textual descriptions.

\subsection{MLLM Interpreter}
\label{sec: MLLM Interpreter}

\subsubsection {Conditional Declaration Language (CDL)}
\noindent \textbf{Definition of CDL.}
We choose Conditional Declaration Language (CDL)~\cite{formalgeo} as the geometric description.
CDL consists of construction statements (ConsCDL) and condition statements (ImgCDL and TextCDL):
ConsCDL defines the fundamental structure of the diagram, \emph{e.g.,} basic shapes, collinear, and cocircular.
ImgCDL and TextCDL specify the geometric and algebraic relations derived from the diagram (ImgCDL) and the problem text (TextCDL), such as the length of line segment or parallel relationships.
Based on topological mapping, CDL offers a unified and rigorous framework for representing plane geometric inputs.

\noindent \textbf{Conciseness of CDL.}
Compared with general textual descriptions, CDL offers a more structured and concise representation (
the proof for CDL's conciseness can be found in the appendix).
Such conciseness inherently narrows the search space and we empirically find that this conciseness benefits the training of MLLM Interpreter (see Sec.~\ref{sec:ablation} for more details) .
%\textcolor{red}{Experimental results are shown in Sec.~\ref{sec:ablation}.}

\subsubsection {Formalgeo7k-Rec-CoT}
Formalgeo7k v2~\cite{FGeo-Parser} is a dataset containing multiple question-answer pairs equipped with corresponding diagrams and CDL annotations.
Based on Formalgeo7k v2, we construct Formalgeo7k-Rec-CoT for training.

\noindent \textbf{Manual Review.}
We identify data quality issues in Formalgeo7k v2, including diagram-question mismatching and incorrect CDL annotations.
To improve the quality of data, we conduct a rigorous manual review.
The review process is completed by four qualified annotators. %with at least an undergraduate-level background.
To ensure objectiveness and consistency, each piece of data is independently reviewed by two annotators.
A modification is applied when a consensus is reached by both annotators; any discrepancy will be adjudicated by the third annotator.

\noindent \textbf{Chain-of-Thought (CoT) Incorporation.}
To guide the MLLM Interpreter to perform step-by-step reasoning for CDL generation, we incorporate Chain-of-Thought (CoT) into Formalgeo7k-Rec-CoT.
The data is structured with special tokens to separate the reasoning process from the CDL output, following the format: $<$think$>$ CoT here $<$/think$>$ $<$cdl$>$ CDL here $<$/cdl$>$.
To automatically generate CoT sequences, we develop a Python parser that extracts and processes geometric information from CDL annotations.
As shown in Fig.~\ref{fig：method}(a), the parser translates the symbolic relations into natural language reasoning steps, creating training pairs for the Supervised Fine-Tuning (SFT) stage.

\subsubsection{Stage 1: SFT with CoT-Augmented Data}
We perform supervised fine-tuning (SFT) on the MLLM Interpreter (e.g., Qwen 2.5-VL and Qwen 3-VL series) using the CoT-augmented training data in Formalgeo7k-Rec-CoT.
The training objective is to minimize the negative log-likelihood of the reasoning steps and final textual description output:
\begin{equation}
    \min_{\theta} \mathbb{E}_{(x,r,y)\sim\mathcal{D}}[-\log\pi_\theta(r,y|x)]
\end{equation}
where $\mathcal{D}$ denotes our training data containing textual description annotations paired with reasoning steps, $x$ denotes the input,  %prompts, 
$r$ denotes the reasoning process, $y$ denotes the final textual description output, and $\pi_{\theta}$ is the probability distribution of the MLLM Interpreter parameterized by $\theta$.
The SFT stage serves as a warm-up phase, equipping the MLLM Interpreter with step-by-step reasoning capability for CDL generation.
However, we observe that SFT alone yields suboptimal performance.
We hypothesize that the standard next-token prediction objective of SFT may inadequately capture the precise logical constraints.
To address the limitation, we explore Group Relative Policy Optimization (GRPO) with CDL matching rewards.

\subsubsection{Stage 2: GRPO with CDL Matching Rewards}
% 改一下题目，

\noindent \textbf{Group Relative Policy Optimization (GRPO).}
During the Reinforcement Learning (RL) stage, we adopt the Group Relative Policy Optimization (GRPO) algorithm~\cite{GRPO}, which optimizes the policy by directly sampling and comparing groups of candidate responses.
Given an input $x$, GRPO samples $n$ candidate responses $\{(r_1, y_1), (r_2, y_2),..., (r_n, y_n)\}$ from the old policy $\pi_{old}$ and evaluates each response by a reward function, producing a set of scores $\{s_1,s_2,...,s_n\}$.
To determine the relative quality of these responses, GRPO normalizes the scores by computing their mean and standard deviation, and then calculates the advantage for each candidate:
\begin{equation}
    A_i = \frac{s_i - \text{mean}(s_1,s_2,...,s_n)}
    {\text{std}(s_1,s_2,...,s_n)},
\end{equation}
where $A_i$ quantifies the advantage of the response $(r_i, y_i)$ relative to other sampled responses. 
The policy $\pi_\theta$ is updated by maximizing the following objective function, which promotes responses with higher advantages while maintaining stability via a KL divergence penalty $\mathbb{D}_{KL}$:
\begin{equation}
\begin{split}
    % \mathcal{J}_{GRPO}(\theta) = \\
    \mathcal{J}(\theta) =
    % \mathbb{E}[{(r_i, y_i)}_{i=1}^n\sim \pi_{old}(x)] \\
\frac{1}{N}\sum_{i=1}^n\{\text{min}[
o_1\cdot A_i,o_2
\cdot A_i]\} 
- \beta\mathbb{D}_{KL}[\pi_\theta||\pi_{ref}],   
\end{split}
\end{equation}
\begin{equation}
    o_1 = \frac{\pi_\theta((r_i,y_i)|x)}{\pi_{\theta_{old}} ((r_i,y_i)|x)},
\end{equation}
\begin{equation}
    o_2 = \text{clip}(
    \frac{\pi_\theta((r_i,y_i)|x)}{\pi_{\theta_{old}} ((r_i,y_i)|x)}
    ,1+\epsilon,1-\epsilon),
\end{equation}
where $\pi_{ref}$ denote the policy of reference model and $\epsilon$, $\beta$ are constants.

\noindent \textbf{Rewards.}
We design specialized rewards in this stage, including CDL matching rewards $S_{{C}}$, $S_{{I}}$, $S_{{T}}$, and Format Reward $S_{{f}}$.
Conventional methods rely on an external LLM solver to derive a solution-based reward.
However, such a solution-based reward is a holistic measure that evaluates the entire set of CDL pieces.
This signal provides sparse guidance, especially when training data is limited and the sampling space is constrained, leading to ineffective optimization.
Instead, we employ CDL matching rewards to provide direct and denser supervision for the generated descriptions.
%\textcolor{red}{Results in Table~\ref{tab:rl_position} validate the superiority of our CDL matching reward design.}

% matching rule
\textbf{(1) CDL Matching Rewards}.
We design matching rewards $S_{{C}}$, $S_{{I}}$, $S_{{T}}$ for ConsCDL, ImgCDL, and TextCDL, respectively.
With matching rewards, 
we aim to encourage the completeness of CDL generation (\emph{i.e.,} high recall) and punish inaccuracy and repetitions (\emph{i.e.,} high precision).
%we combine the recall and precision of generated CDL to encourage the completeness and preciseness of generation, while punishing inaccuracy or repetitions by the precision score.
Thus, we choose to evaluate the recall and precision of CDL generation results by performing greedy matching.

We take the matching reward for ConsCDL $S_{{C}}$ as an example.
We implement a Python script to judge the matching of ConsCDL, \emph{e.g.}, two pieces that refer to the same geometric shape are considered a matched pair.
Formally, let $G = \{g_1, g_2, \dots, g_m\}$ be the ground truth set, while $P = \{p_1, p_2, \dots, p_n\}$ be the predicted set.
The algorithm iterates through each predicted item $p \in P$, and searches for the first item in the current ground truth set $g \in G$ that exactly matches $p$.
Once finding a match, the corresponding ground truth item $g$ is removed from $G$.
This process continues until all items in the predicted set $P$ are evaluated or the ground truth set becomes empty.

We denote the remaining set of unmatched ground truth items after this procedure as $G'$, so the number of successfully matched pairs is then $M = m - |G'|$.
We measure the \textit{recall} score $R_{\text{ConsCDL}}$ to encourage completeness of ConsCDL generation, and measure the \textit{precision} score $A_{\text{ConsCDL}}$ to penalize inaccurate or repeated pieces.
The final reward $S_{{C}}$ is defined as the average of the precision score and the recall score
\begin{equation}
    R_{\text{ConsCDL}} = \frac{M}{m}, P_{\text{ConsCDL}} = \frac{M}{n},
\end{equation}
\begin{equation}
    S_{{C}} = \frac{R_{\text{ConsCDL}} + P_{\text{ConsCDL}}}{2}.
\end{equation}
Matching reward for ImgCDL $S_{{I}}$ and TextCDL $S_{{T}}$ are calculated in the same way as that for $S_{{C}}$.

\textbf{(2) Format Reward $S_{{f}}$} 
We also design a format reward $S_{{f}}$ to evaluate whether the model's output strictly follows the format that requires the model to output special tags (\emph{e.g.}, $<$think$>$ ... $<$/think$>$ $<$cdl$>$ ... $<$/cdl$>$).
The reward is set to $1$ if the output complies with the formatting rules; otherwise, the reward is $0$.

The overall reward is calculated as 
\begin{equation}
%\begin{split}
    S = \alpha S_{{f}} + \gamma S_{{C}} + (1-\alpha-\gamma)(S_{{I}}+S_{{T}}),
%\end{split}
\end{equation}
where $\alpha$ and $\gamma$ are constants set to $0.1$ and $0.5$. %respectively.

\subsection{LLM Solver with Generated CDL}
\label{sec: LLM Solver}

We adopt an off-the-shelf LLM (\emph{e.g.,} Qwen3 30B~\cite{Qwen3}) to finalize the PGPS task. Specifically, LLM solver utilizes the generated CDL from the MLLM Interpreter without the CoT as input.
Based on the generated CDL, the LLM Solver performs reasoning and derives the final answer.
%The overall performance is evaluated by the problem solving accuracy across multiple PGPS benchmarks.

\begin{table*}
\setlength{\tabcolsep}{6.3pt}
\renewcommand{\arraystretch}{1.4}
  \caption{\textbf{Comparison of CDL Generation performance on Formalgeo-Rec-CoT.} 
  Fgeo-Parser~\cite{FGeo-Parser} employs a diagram parser (denoted as ``(Diag.)'') to convert geometric diagrams into ConsCDL and ImgCDL, and a text parser (denoted as ``(Text)'') to transform problem textural description into textCDL.
  Diagram Formalizer~\cite{Diagram-Formalizer} is designed solely for parsing geometric diagrams into ConsCDL and ImgCDL.
  Our proposed MLLM Interpreter can process both diagrams and text with a unified model and exhibits strong overall performance.}
  \label{tab:cdl_comparison}
  \centering
  \begin{tabular}{c|c|c|cc|cc|cc}
    \toprule[1pt]
    \multirow{2}{*}{\makecell[c]{Methods}} & \multirow{2}{*}{\makecell[c]{Pub. \& Year}} & \multirow{2}{*}{\makecell[c]{Training\\Data}} &\multicolumn{2}{c|}{TextCDL} & \multicolumn{2}{c|}{ImgCDL} & \multicolumn{2}{c}{ConsCDL} \\
     & & & Recall & Precision & Recall & Precision & Recall & Precision\\
    \midrule
    FgeoParser (Diag.)~\cite{FGeo-Parser} & Symmetry'24 &\multirow{2}{*}{\makecell[c]{$14.7$k}} & - & - & $77.5$ & - & $87.0$ & -\\
    FgeoParser (Text)~\cite{FGeo-Parser} & Symmetry'24 & & $96.5$ & - & - & - & - & -\\
    \midrule
    Diagram Formalizer~\cite{Diagram-Formalizer} & ICASSP'25 & $1$M & - & - & $92.9$ & - & $90.3$ & - \\
    \midrule
    Qwen2.5-VL 3B (SFT) & \multirow{4}{*}{\makecell[c]{Ours}} & \multirow{4}{*}{\makecell[c]{$5.5$k}}& $94.7$ & $94.7$ & $94.7$ & $92.2$ & $62.3$ & $58.7$\\
    Qwen2.5-VL 3B (RL) & & & $98.9$ & $98.9$ & $96.5$ & $96.9$ & $87.1$ & $87.6$\\
    Qwen2.5-VL 7B (SFT) & & & $96.5$ & $96.6$ & $95.1$ & $93.5$ & $70.0$ & $68.3$\\
    Qwen2.5-VL 7B (RL) & & & \textbf{99.1} & \textbf{99.1} & \textbf{97.0} & \textbf{96.9} & \textbf{92.7} & \textbf{92.1}\\
    \midrule
    Qwen3-VL 4B (SFT) & \multirow{4}{*}{\makecell[c]{Ours}} & \multirow{4}{*}{\makecell[c]{$5.5$k}} & $97.1$ & $97.1$ & $96.4$ & $95.1$ & $80.9$ & $79.4$ \\
    Qwen3-VL 4B (RL) & & &\textbf{99.1} & \textbf{99.1} & \textbf{97.2} & \textbf{97.2} & \textbf{95.4} & \textbf{95.1}\\
    Qwen3-VL 8B (SFT)& & &$97.2$ & $97.2$ & $96.0$ & $95.0$ & $80.9$ & $79.6$ \\
    Qwen3-VL 8B (RL) & & & \textbf{99.2} & \textbf{99.3} & \textbf{97.0} & \textbf{96.9} & \textbf{95.9} & \textbf{95.8}\\
    \bottomrule[1pt]
  \end{tabular}
\end{table*}

\begin{table*}
\setlength{\tabcolsep}{7pt}
\renewcommand{\arraystretch}{1.4}
  \caption{\textbf{Effect of CoT in CDL generation.}
  We perform the experiments on Qwen2.5-VL 7B.
  Results show that incorporating CoT of ConsCDL into the training data boosts both CDL generation performance and solving accuracy. CoTs of ImgCDL and TextCDL introduce extra tokens but degrade the performance.
  Considering both training cost and performance, we only utilize CoT of ConsCDL.}
  \label{tab:ablation_cot}
  \centering
  \begin{tabular}{ccc|cc|cc|cc|c}
    \toprule[1pt]
    \multicolumn{3}{c|}{{CoT}} & \multicolumn{2}{c|}{TextCDL} & \multicolumn{2}{c|}{ImgCDL} & \multicolumn{2}{c|}{ConsCDL} & \multirow{2}{*}{\makecell[c]{Formalgeo}} \\
     ConsCDL& ImgCDL& TextCDL & Recall & Precision & Recall & Precision & Recall & Precision\\
    \midrule
    & & & \textbf{99.1}& $98.4$ & \textbf{97.0}& $96.9$ &$81.7$& $81.5$ & $76.3$\\
    \checkmark & & & \textbf{99.1} & \textbf{99.1} & \textbf{97.0} & $96.9$ & \textbf{92.7} & \textbf{92.1} & \textbf{83.2}\\
    \checkmark & \checkmark & & $98.5$& $98.4$ & \textbf{97.0}& \textbf{97.1} & $92.4$ & $92.0$ & $82.8$\\
    \checkmark & & \checkmark & \textbf{99.1} & \textbf{99.1} & $96.9$ & $96.9$ & $92.0$  & $91.7$ & $81.7$\\
    \bottomrule[1pt]
  \end{tabular}
\end{table*}

\section{Experiment}
\label{4_exp}

\begin{table}
\setlength{\tabcolsep}{2pt}
\renewcommand{\arraystretch}{1.4}
  \caption{\textbf{Comparison of Plane Geometry Problem Solving performance with closed-source and open-source MLLMs.} 
  On both in-domain and out-of-domain benchmarks, our proposed method significantly outperforms all open-source MLLMs and shows comparable performance to SOTA closed-source MLLMs.}
  \label{tab:pgps}
  \centering
  \begin{tabular}{l|c|c|cc}
    \toprule[1pt]
    % \multirow{2}{*}{\makecell[c]{Models}} & \multirow{2}{*}{\makecell[c]{Training\\Data}} & {In Domain} & \multicolumn{2}{c}{OOD} \\
    Models & Data &Formalgeo & Unigeo & MathV. \\
    \midrule
    \multicolumn{4}{c}{\textbf{Closed-Source MLLMs}}\\
    \midrule
    GPT-4o & \multirow{6}{*}{\makecell[c]{-}} & $58.0$ & $43.9$ & $47.1$\\
    Claude-Sonnet-4 & & $69.1$ & $72.6$ & $60.7$\\
    Claude-Opus-4.1 & & $69.1$ & $74.2$ & $63.1$\\
    Gemini2.5-Flash & & $80.5$ & $82.8$ & $79.4$\\
    Gemini2.5-Pro & & $81.8$ & \textbf{84.6} & \textbf{81.3} \\
    % GPT-5 & & \textbf{86.7} & \textbf{87.0} & \textbf{89.3} \\
    \midrule
    \multicolumn{4}{c}{\textbf{Open-Source MLLMs}}\\
    \midrule
    G-LLaVA~\cite{G-LLaVA} & $170$k & - & - & $56.7$\\
    Qwen2.5-VL 32B~\cite{Qwen2.5-vl} & -& $57.3$ & $67.9$ & $66.8$\\
    % Qwen3-VL 30B & - & $80.1$ & \textbf{86.3}  & $80.8$\\
    GLM4.1-V~\cite{GLM-V} &-& $73.4$ & $79.2$ & $80.4$ \\
    GeoUni~\cite{geouni} &$235$k& $59.8$ & - & -\\
    DFE-GPS~\cite{Diagram-Formalizer} & $238$k & $75.3$ & -& -\\
    GF Reasoner~\cite{GFReasoner} & $184$k & - &$72.7$ & $64.9$\\
    \midrule
     \multicolumn{2}{c|}{}& \textbf{In Domain} & \multicolumn{2}{c}{\textbf{OOD}}\\
    \midrule
    Ours & $5.5$k & \textbf{85.7} & \textbf{84.0} & \textbf{80.8}\\
    \bottomrule[1pt]
  \end{tabular}
\end{table}

\subsection{Setup}
\noindent \textbf{Base Model.}
We conduct experiments with Qwen 2.5-VL~\cite{Qwen2.5-vl} series and Qwen 3-VL series as the MLLM Interpreter.
For the LLM solver, we adopt Qwen3 30B~\cite{Qwen3} for its strong logical reasoning capability.

\noindent \textbf{Training Dataset.}
We utilize the constructed Formalgeo7k-Rec-CoT as our training dataset.
Originally, Formalgeo7k v2~\cite{FGeo-Parser} contains $7,000$ geometric question-answer pairs, each equipped with a diagram and corresponding CDL annotation.
We construct Formalgeo7k-Rec-CoT by conducting a rigorous manual review on Formalgeo7k v2 and incorporating Chain-of-Thoughts (CoTs).
Then, we randomly divide Formalgeo7k-Rec-CoT into training and validation sets with a ratio of $0.8$ and $0.2$, resulting in $5,550$ pairs for training and $1,390$ pairs for validation.

\noindent \textbf{Implementation details.}
For the MLLM Interpreter, the SFT stage runs for $3$ epochs with batch size $8$ and learning rate $1e-5$ .
The RL stage lasts for $15$ epochs with batch size $128$, learning rate $1e-6$, and a rollout number $N=8$.
The LLM solver is always kept frozen.

\noindent \textbf{Evaluation Benchmarks.}
Firstly, we evaluate the CDL generation performance of the MLLM Interpreter on the validation set of Formalgeo7k-Rec-CoT.
Besides, to evaluate the plane geometric problem-solving performance of our method, we utilize both in-domain benchmark Formalgeo7k-Rec-CoT and out-of-domain benchmarks, including Unigeo~\cite{Unigeo} and the geometric set of MathVista~\cite{mathvista}.
All evaluations are open-ended: we extract the final answer from the LLM solver's response and compare it with the ground truth to judge correctness.

\subsection{Quantitative Results}
\subsubsection{CDL Generation}
\noindent \textbf{Baseline.}
We compare the CDL Generation against two previous models, Fgeo-Parser~\cite{FGeo-Parser} and Diagram Formalizer~\cite{Diagram-Formalizer}, on the validation set of Formalgeo7k-Rec-CoT.
Fgeo-Parser~\cite{FGeo-Parser} employs two separate parsers: a diagram parser based on BLIP~\cite{blip} to convert geometric diagrams into ConsCDL and imgCDL, and a text parser based on T5~\cite{T5} to transform problem textural description into textCDL.
Both parsers are trained on an augmented version of the Formalgeo7k v2 training set with $14,700$ samples.
Diagram Formalizer~\cite{Diagram-Formalizer} is trained on large-scale datasets, including formalgeo-structure774k~\cite{Diagram-Formalizer} and SynthGeo228k~\cite{Diagram-Formalizer}, and is designed solely for parsing geometric diagrams into ConsCDL and ImgCDL.
In contrast, our proposed MLLM Interpreter can process both diagrams and text with a unified model, while requiring only $5.5k$ training samples.

\noindent \textbf{Comparison. }
Consistent with CDL matching rewards in GRPO, we evaluate CDL generation performance using recall and precision scores (see Eq.~(6)) for each CDL type.
As shown in Table~\ref{tab:cdl_comparison}, our MLLM Interpreter exhibits strong overall performance on CDL generation.
% 评价指标

\subsubsection{Plane Geometry Problem Solving}
\noindent \textbf{Baseline.}
We compare the plane geometry problem solving performance of our method with both closed-source and open-source leading MLLMs.
The closed-source baselines include the available state-of-the-art (SOTA) models: GPT-4o~\cite{Gpt-4o}, Claude series, and Gemini series~\cite{Gemini2.5}.
For open-source MLLMs, we evaluate general-purpose reasoning models such as Qwen2.5-VL 32B~\cite{Qwen2.5-vl} and GLM 4.1-V-Thinking~\cite{GLM-V}, as well as previous methods specifically designed for PGPS, including G-LLaVA~\cite{G-LLaVA}, GeoUni~\cite{geouni}, DFE-GPS~\cite{Diagram-Formalizer}, and GF Reasoner~\cite{GFReasoner}.

\noindent \textbf{Comparison.}
We perform open-ended evaluation and report the solving accuracy across benchmarks.
As shown in Table~\ref{tab:pgps}, our method consistently achieves superior performance on both in-domain and out-of-domain benchmarks.
With only $5.5$k training data, it significantly outperforms all open-source MLLMs and attains comparable performance to state-of-the-art closed-source model Gemini2.5-Pro.
Notably, on Formalgeo-Rec-CoT, we surpass Gemini2.5-Pro by $3.9\%$ accuracy.

\subsection{Ablation Study}
\label{sec:ablation}
In this section, all ablation experiments are performed with Qwen2.5-VL 7B as the MLLM Interpreter.

\noindent \textbf{Concise descriptions benefit MLLM's learning by narrowing the search space. }
To empirically validate that the conciseness of CDL benefits the MLLM Interpreter's training, we conduct an ablation study (Table~\ref{tab:ablation_concisness}).
We intentionally introduce redundant information into the ConsCDL annotations (Line 1), \emph{e.g.}, replacing preset terms like ``Shape'' with more specific but unnecessary ones such as ``Triangle'' and ``Circle Segment''.
This intentional expansion of the description search space leads to a degradation in ConsCDL generation performance.
Consequently, compared to generating the original, concise CDL (Line 2), the expanded version results in lower solving accuracy on Formalgeo7k-Rec-CoT, thereby directly confirming the benefit of a narrowed search space.

\noindent \textbf{CDL Matching Rewards outperform Solution-based Reward. }
Table~\ref{tab:ablation_solution} compares our designed CDL matching rewards to conventional solution-based reward in GRPO. 
Combining solution-based reward with ours, the performance drops obviously.
With solution-based reward only, there is no performance gain compared with SFT.
All those results verify that CDL matching rewards are more effective than solution-based reward.
%We conduct an ablation in Table~\ref{tab:ablation_solution} to validate that, compared with solution-based rewards, CDL Matching Rewards provide more direct and dense guidance for CDL generation.
%According to the results, introducing an extra solution-based reward degrades the CDL generation performance, leading to a degradation in solving accuracy. 
%Consequently, CDL Matching Rewards outperform solution-based rewards during the GRPO stage.

\begin{table}
\setlength{\tabcolsep}{4.3pt}
\renewcommand{\arraystretch}{1.3}
  \caption{\textbf{Concise descriptions benefit MLLM Interpreter's training by narrowing search space.} 
  We conduct an ablation to validate that the conciseness of CDL benefits the learning of the MLLM Interpreter (Qwen2.5-VL 7B) on Formalgeo-Rec-CoT.}
  % 训练结果
  \label{tab:ablation_concisness}
  \centering
  \begin{tabular}{c|cc|cc|cc|c}
    \toprule[1pt]
    \multirow{2}{*}{\makecell[c]{Concise}}  &\multicolumn{2}{c|}{TextCDL} & \multicolumn{2}{c|}{ImgCDL} & \multicolumn{2}{c|}{ConsCDL} & \multirow{2}{*}{\makecell[c]{Acc.}} \\
    & Re. & Pre. & Re. & Pre. & Re. & Pre.\\
    \midrule
     & \textbf{99.1} & \textbf{99.1} & \textbf{97.0} & \textbf{97.0} & $91.6$ & $90.7$ & $80.5$\\
    \checkmark & \textbf{99.1} & \textbf{99.1} & \textbf{97.0} & $96.9$ & \textbf{92.7} & \textbf{92.1} & \textbf{83.2}\\
    \bottomrule[1pt]

  \end{tabular}
\end{table}

\begin{table}
\setlength{\tabcolsep}{2.8pt}
\renewcommand{\arraystretch}{1.3}
  \caption{\textbf{CDL Matching Rewards outperform the Solution-based Reward.} 
  We train the MLLM Interpreter (Qwen2.5-VL 7B) with an extra solution-based reward provided by LLM Solver (Qwen3 30B) and observe that the sparse reward degrades the performance on Formalgeo-Rec-CoT.}
  \label{tab:ablation_solution}
  \centering
  \begin{tabular}{cc|cc|cc|cc|c}
    \toprule[1pt]
    \multirow{2}{*}{\makecell[c]{Ours}} & \multirow{2}{*}{\makecell[c]{Solution \\ Reward}}  &\multicolumn{2}{c|}{TextCDL} & \multicolumn{2}{c|}{ImgCDL} & \multicolumn{2}{c|}{ConsCDL} & \multirow{2}{*}{\makecell[c]{Acc.}} \\
     & & Re. & Pre. & Re. & Pre. & Re. & Pre.\\
    \midrule
    \checkmark & &\textbf{99.1} & \textbf{99.1} & \textbf{97.0} & $96.9$ & \textbf{92.7} & \textbf{92.1} & \textbf{83.2} \\
     % & \checkmark & low & low & low & low & low & low & low\\
    % & \checkmark\\
    \checkmark & \checkmark & $98.9$ & $99.0$ & $96.4$ & $96.5$ & $91.5$ & $91.1$ & $81.3$\\
    \bottomrule[1pt]

  \end{tabular}
\end{table}
\begin{table*}
\setlength{\tabcolsep}{9pt}
\renewcommand{\arraystretch}{1.4}
  \caption{\textbf{Effect of rewards in RL stage.}
  In order to validate the effect of rewards in the RL stage, we perform the ablations with Qwen2.5-VL 7B on Formalgeo-Rec-CoT.
  Results further validate that each proposed reward contributes to a better performance.}
  \label{tab:ablation_reward}
  \centering
  \begin{tabular}{cccc|cc|cc|cc|c}
    \toprule[1pt]
    \multicolumn{4}{c|}{Rewards} & \multicolumn{2}{c|}{TextCDL} & \multicolumn{2}{c|}{ImgCDL} & \multicolumn{2}{c|}{ConsCDL} & \multirow{2}{*}{\makecell[c]{Formalgeo}} \\
    $S_C$ & $S_I$ & $S_T$ & $S_f$& Recall & Precision & Recall & Precision & Recall & Precision\\
    \midrule
    \checkmark & & & & $95.9$ & $95.8$ & $95.7$ & $94.1$ & $92.0$ & $91.8$ & $81.0$\\
    \checkmark & \checkmark & & & $96.5$ & $96.5$ & \textbf{97.1} & \textbf{97.1} & $92.5$ & $91.7$ & $82.2$\\
    \checkmark & \checkmark & \checkmark & & $98.3$ & $98.3$ & $97.0$ & $97.0$ & $92.0$ & $91.5$ & $81.6$\\
    \checkmark & \checkmark & \checkmark & \checkmark & \textbf{99.1} & \textbf{99.1} & $97.0$ & $96.9$ & \textbf{92.7} & \textbf{92.1} & \textbf{83.2}\\
    \bottomrule[1pt]

  \end{tabular}
\end{table*}

\begin{figure*}
  \centering
  \includegraphics[width=0.99\linewidth]{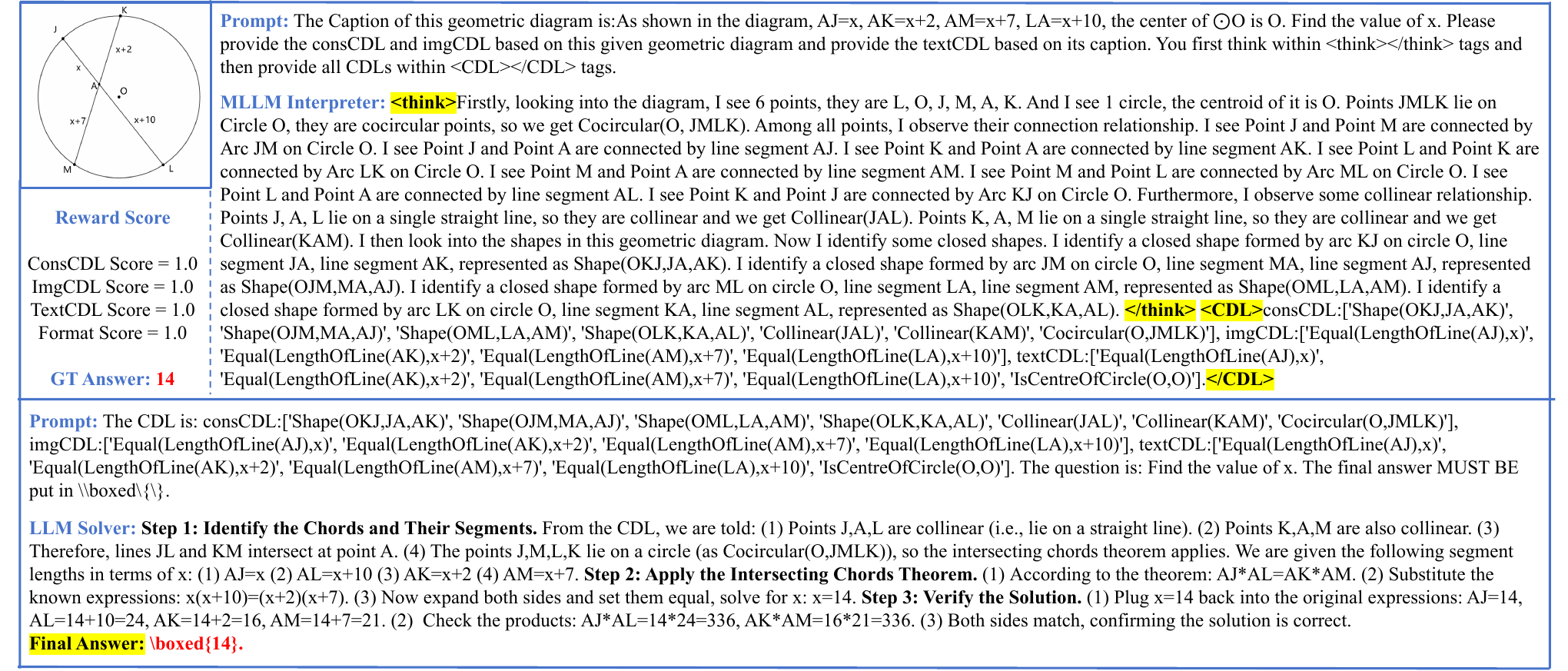}
  \caption{
  \textbf{Qualitative Results on Formalgeo7k-Rec-CoT.} We present an example illustrating the complete pipeline of CDL Solver: an MLLM Interpreter converts geometric inputs into CDL upon which the LLM Solver performs reasoning and derives the final answer.}
\label{fig：visualization}
\end{figure*}

\noindent \textbf{Effect of CoT in CDL generation.}
During the SFT stage, we augment the training data solely with ConsCDL CoT, excluding CoTs for ImgCDL and TextCDL.
As evidenced by Table~\ref{tab:ablation_cot}, training without CoT (row 1) severely degrades CDL quality (especially ConsCDL) and yields low solving accuracy.
Introducing ConsCDL CoT (row 2) yields consistent performance improvements.
However, incorporating CoTs for ImgCDL and TextCDL leads to a slight degradation.
We hypothesize that generating ImgCDL and TextCDL relies more on direct perception from the diagram or caption, requiring less step-by-step reasoning than ConsCDL.
Consequently, the additional CoT tokens may act as a distraction, which increases the training complexity without benefit.
Considering both efficiency and performance, we only incorporate CoT of ConsCDL into training data.

\noindent \textbf{Effect of various RL rewards: $S_f$, $S_C$, $S_I$ and $S_T$. }
During the RL stage, we specifically design rewards, including Format Reward $S_f$, ConsCDL Reward $S_C$, ImgCDL Reward $S_I$, and TextCDL Reward $S_T$.
In order to validate the effect of each reward, we design ablations in Table~\ref{tab:ablation_reward}.
Results show that each reward contributes to the CDL generation and geometric problem solving performance.

\subsection{Qualitative Results}
% 放到最后
We present an example on the validation set of Formalgeo7k-Rec-CoT to illustrate the complete pipeline of our proposed CDL Solver: the MLLM Interpreter converts the diagrams and captions into CDL upon which the LLM Solver performs reasoning and derives the final answer.
More qualitative results can be found in the appendix.

\section{Conclusion}
In this paper, we observe that LLM itself is potentially a powerful PGPS solver when converting visual diagram into textual descriptions.
Thus, we propose a new paradigm to unleash LLM's potential for Plane Geometry Problem Solving (PGPS).
We train an MLLM Interpreter to convert geometric diagram into a concise geometric description (CDL) and then an off-the-shelf LLM Solver is utilized to perform reasoning.
We design a two-stage pipeline for the training of MLLM Interpreter, including an SFT Stage with CoT-augmented data and a GRPO Stage with specifically designed CDL matching rewards.
Extensive experiments demonstrate that our method achieves superior performance in both CDL generation and solving accuracy on PGPS benchmarks.
We hope that our work may inspire future research to investigate how to better exploit the inherent reasoning capability of LLM for complex multimodal reasoning tasks.

\clearpage

{
    \small
    \bibliographystyle{ieeenat_fullname}
    \bibliography{main}
}

\clearpage
\setcounter{page}{1}
\maketitlesupplementary

\section{More Ablations}
\label{sec:sup_ablation}
\noindent \textbf{Effect of rollout number $N$ in GRPO. }
In order to validate the effect of rollout number $N$ in the GRPO stage, we perform an ablation study on Qwen2.5-VL 7B in Table~\ref{tab:ablation_rollout}.
Setting $N=10$ yields no performance gain on CDL generation and slightly degrades the problem solving accuracy. Moreover, it brings an extra $80$ hours of training time compared with $N=8$.
Considering the training cost and the performance, we set $N=8$.
\begin{table}
\setlength{\tabcolsep}{7pt}
\renewcommand{\arraystretch}{1.4}
  \caption{\textbf{Effect of rollout number $N$ in the GRPO stage.} 
  We perform the experiments on Qwen2.5-VL 7B.
  Accounting for both training cost and performance, we set $N=8$.
  }
  \label{tab:ablation_rollout}
  \centering
  \begin{tabular}{c|c|c|c|c}
    \toprule[1pt]
    \multicolumn{2}{c|}{Performance} & $N=5$ & {$N=\bf{8}$} & $N=10$\\
    \midrule
    \multirow{2}{*}{\makecell[l]{TextCDL}} & Recall & $98.5$ & \textbf{99.1} & $98.3$\\
     & Precision & $98.4$ & \textbf{99.1} & $98.3$\\
     \midrule
    \multirow{2}{*}{\makecell[l]{ImgCDL}} & Recall & $96.9$ & \textbf{97.0} & \textbf{97.0}\\
     & Precision & $96.9$ & $96.9$ & \textbf{97.0}\\
     \midrule
    \multirow{2}{*}{\makecell[l]{ConsCDL}} & Recall & $89.1$ & \textbf{92.7} & $91.0$\\
     & Precision & $87.9$ & \textbf{92.1} & $90.3$\\
    \midrule
    \multicolumn{2}{c|}{Formalgeo-Rec-CoT} & $80.9$ & \textbf{83.2} & $82.5$\\
    \bottomrule[1pt]
  \end{tabular}
\end{table}

\noindent \textbf{Effect of Reward Weights  $\alpha$, $\gamma$ in GRPO. }
In order to validate the effect of reward weights $\alpha$, $\gamma$ in the GRPO stage, we perform an ablation study on Qwen3-VL 8B in Table~\ref{tab:ablation_reward_weight}.
Considering the CDL generation performance and the solving accuracy, we set $\alpha=0.1, \gamma=0.5$.

\begin{table}
\setlength{\tabcolsep}{4.5pt}
\renewcommand{\arraystretch}{1.3}
  \caption{\textbf{Effect of Reward Weights  $\alpha$, $\gamma$ in GRPO.} 
  We perform the experiments on Qwen3-VL 8B.
  According to results, we set $\alpha=0.1, \gamma=0.5$.}
  \label{tab:ablation_reward_weight}
  \centering
  \begin{tabular}{cc|cc|cc|cc|c}
    \toprule[1pt]
    \multirow{2}{*}{\makecell[c]{$\alpha$}} & \multirow{2}{*}{\makecell[c]{$\gamma$}} &\multicolumn{2}{c|}{TextCDL} & \multicolumn{2}{c|}{ImgCDL} & \multicolumn{2}{c}{ConsCDL} & \multirow{2}{*}{\makecell[c]{Acc.}}\\
    & & Re. & Pre. & Re. & Pre. & Re. & Pre.\\
    \midrule
     $0.2$ & $0.4$ & $98.3$ & $98.9$ & $96.8$ & $96.8$ & \textbf{96.1} & \textbf{96.0} & $84.4$\\
     $0.4$ & $0.3$ & $98.4$ & $99.0$ & \textbf{97.2} & \textbf{97.0} & \textbf{96.1} & \textbf{96.0} &  $83.6$\\
     $0.1$ & $0.5$ & \textbf{99.2} & \textbf{99.3} & $97.0$ & $96.9$ & $95.9$ & $95.8$ & \textbf{85.7}\\
    \bottomrule[1pt]

  \end{tabular}
\end{table}

\noindent \textbf{Effect of learning rate in SFT and RL stages.}
During the training process, we adopt the learning rate of $1.0\times e^{-5}$ for SFT and $1.0\times e^{-6}$ for RL.
We further conduct ablations on various choices of learning rate.
Results confirm that the chosen hyperparameters yield the best results, and the performance remains relatively stable across various choices.

\noindent \textbf{Superior Performance on Original Formalgeo7k v2.}
In Table~\ref{tab:ablation_review}, we conduct both training and evaluation on the original Formalgeo7k v2 (``Ours (Formalgeo7k)'' in the table).
The results demonstrate that our method outperforms previous CDL generation approaches even when using the original data, validating the effectiveness of our framework.
Furthermore, training with the refined data yields better performance, confirming that high-quality CDL annotations are also beneficial.

\begin{table}
\setlength{\tabcolsep}{3pt}
\renewcommand{\arraystretch}{1.6}
  \caption{\textbf{Superior Performance on Original Formalgeo7k v2.} 
  We further conduct both training and evaluation on the original Formalgeo7k v2.
  Compared with previous methods, our superior performance on the original Formalgeo7k v2 validates the effectiveness of our framework.}
  \label{tab:ablation_review}
  \centering
  \begin{tabular}{c|cc|cc|cc}
    \toprule[1pt]
    \multirow{2}{*}{\makecell[c]{Methods}} &\multicolumn{2}{c|}{TextCDL} & \multicolumn{2}{c|}{ImgCDL} & \multicolumn{2}{c}{ConsCDL}\\
     & Re. & Pre. & Re. & Pre. & Re. & Pre.\\
    \midrule
    FgeoParser (Diag.) & - & - & $77.5$ & - & $87.0$ & - \\
    FgeoParser (Text) & $96.5$ & - & - & - & - & -\\
    \midrule
    Diagram Formalizer & - & - & $92.9$ & - & $90.3$ & -\\
    \midrule
    Ours (Formalgeo7k) & $98.2$ & $98.3$ & \textbf{97.0} & $97.0$ & $91.0$ & $90.3$\\
    Ours (Rec-CoT)&  \textbf{99.1} & \textbf{99.1} & \textbf{97.0} & \textbf{96.9} & \textbf{92.7} & \textbf{92.1}\\
    \bottomrule[1pt]
  \end{tabular}
\end{table}

\noindent \textbf{Empirical analysis of how concise descriptions benefit MLLM's learning.}
We further compare the performance of concise CDL against the expanded version on the Formalgeo-Rec-CoT training set.
While results in Table~\ref{tab:ablation_concisness_train} show that the expanded version achieves slightly higher CDL matching scores on the training set, its performance on the validation set is inferior to our concise CDL (as shown in Table~\ref{tab:ablation_concisness}).
This phenomenon shows the relatively poor generalization of the expanded version compared to ours.
The expanded version introduces unnecessary or redundant information, which unexpectedly enlarges the search space. Consequently, with relatively limited samples, MLLM tends to memorize training samples rather than learning generalizable patterns.
In contrast, the search space of concise CDL is smaller, which eases the generalization.

\begin{table}
\setlength{\tabcolsep}{6.5pt}
\renewcommand{\arraystretch}{1.3}
  \caption{\textbf{Concise descriptions (ours) \emph{v.s.} Expanded version on the training set.} 
  While the expanded version achieves slightly higher scores than the concise CDL on the training set, its performance on the validation set (shown in Table~\ref{tab:ablation_concisness}) degrades, which shows poor generalization.}
  \label{tab:ablation_concisness_train}
  \centering
  \begin{tabular}{c|cc|cc|cc}
    \toprule[1pt]
    \multirow{2}{*}{\makecell[c]{Type}}  &\multicolumn{2}{c|}{TextCDL} & \multicolumn{2}{c|}{ImgCDL} & \multicolumn{2}{c}{ConsCDL} \\
    & Re. & Pre. & Re. & Pre. & Re. & Pre.\\
    \midrule
     Concise & $99.3$ & $99.4$ & \textbf{98.7} & $98.5$ & $93.8$ & \textbf{94.1} \\
     Expanded & \textbf{99.8} & \textbf{99.8} & \textbf{98.7} & \textbf{98.7} & \textbf{94.0} & \textbf{94.1}\\
    \bottomrule[1pt]

  \end{tabular}
\end{table}

\noindent \textbf{Generalization of CDL for various LLM Solvers. }
% Generalization
In order to validate the generalization capability of our proposed diagram, we further perform an ablation for various LLM Solvers.
Taking the generated CDL from our MLLM Interpreter (Qwen2.5-VL 7B) as inputs, we utilize various LLMs to serve as solvers, including Qwen3 30B, Qwen3 32B, DeepSeek-V3.1-Terminus, and GLM-4.6.
Results in Table~\ref{tab:ablation_llm} demonstrate that our proposed paradigm achieves overall superior performance against previous methods across LLM solvers, validating its strong generalization.
Qwen3 30B yields the best result and is therefore selected as the LLM solver in our main paper. 

\begin{table}
\setlength{\tabcolsep}{13pt}
\renewcommand{\arraystretch}{1.4}
  \caption{\textbf{Generalization of CDL for various LLM Solvers.} 
  We perform experiments with various LLM Solvers on Formalgeo-Rec-CoT. 
  All LLM Solvers take the generated CDL from our MLLM Interpreter (Qwen2.5-VL 7B) as inputs to perform reasoning.}
  \label{tab:ablation_llm}
  \centering
  \begin{tabular}{c|c|c}
    \toprule[1pt]
    % \multirow{2}{*}{\makecell[c]{Models}} & \multirow{2}{*}{\makecell[c]{Training\\Data}} & {In Domain} & \multicolumn{2}{c}{OOD} \\
    \multicolumn{2}{c|}{\textbf{Models}} & Formalgeo \\
    \midrule
    \multicolumn{3}{c}{\textbf{Closed-Source MLLMs}}\\
    \midrule
    \multicolumn{2}{c|}{GPT-4o} &  $58.0$\\
    \multicolumn{2}{c|}{Claude-Sonnet-4} &  $69.1$\\
   \multicolumn{2}{c|}{Claude-Opus-4.1}  &  $69.1$\\
    \multicolumn{2}{c|}{Gemini2.5-Flash} &  $80.5$\\
    \multicolumn{2}{c|}{Gemini2.5-Pro}   &  $81.8$\\
    \midrule
    \multicolumn{3}{c}{\textbf{Open-Source MLLMs}}\\
    \midrule
    \multicolumn{2}{c|}{Qwen2.5-VL 32B} &  $57.3$\\
    \multicolumn{2}{c|}{Qwen3-VL 30B} &  $80.1$\\
    \multicolumn{2}{c|}{GLM4.1-V} &  $73.4$\\
    \multicolumn{2}{c|}{GeoUni} &  $59.8$\\
    \multicolumn{2}{c|}{DFE-GPS}&  $75.3$\\
    \midrule
    \multicolumn{3}{c}{\textbf{Ours}}\\
    \midrule
    \multirow{4}{*}{\makecell[c]{MLLM\\Interpreter}}  &  +Qwen3 32B & $82.5$\\
    & +GLM-4.6 & $82.3$\\
    & +DeepSeek V3.1 & $79.4$\\ 
    & +Qwen3 30B & \textbf{83.2}\\
    \bottomrule[1pt]
  \end{tabular}
\end{table}

\noindent \textbf{Solving Accuracy Comparison with Diagram Formalizer on the same LLM Solver. }
We further evaluate end-to-end solving accuracy improvements against DFE-GPS's Diagram Formalizer~\cite{Diagram-Formalizer} across both in-domain (Formalgeo7k v2) and Out-Of-Domain (Unigeo \& Mathvista) benchmarks. 
CDL generation results of Diagram Formalizer are obtained by running its official code.
In Table~\ref{tab:ablation_dfe-gps}, using the \textbf{same LLM solver (Qwen3 30B)}, ours yields consistent improvement across all benchmarks, further validating our generalization capability and confirming that enhancement on CDL generation indeed leads to better solving performance. 
\begin{table}[ht]
\setlength{\tabcolsep}{6pt}
\renewcommand{\arraystretch}{1}
  \caption{\textbf{Solving Accuracy Comparison with Diagram Formalizer on the same LLM Solver.} 
  We evaluate end-to-end solving accuracy improvements against DFE-GPS's Diagram Formalizer on the same LLM Solver, Qwen3 30B.
  Results demonstrate the generalization capability of our method and confirm that enhancement on CDL generation indeed leads to better solving performance. }
  \label{tab:ablation_dfe-gps}
  \centering
  \begin{tabular}{l|c|ccc}
    \midrule
    Models & LLM& Formalgeo & Unigeo & MathV. \\
    \midrule
    % \multirow{2}{*}{\makecell[c]{DFE-GPS}} &Yi1.5-Chat-34B & $56.9$ & $46.6$ & $36.3$\\
    DFE-GPS& Qwen3& $80.3$ & $79.9$ & $69.0$\\
    % \midrule
    Ours & Qwen3&\textbf{85.7} & \textbf{84.0} & \textbf{80.8}\\
    \midrule
  \end{tabular}
\end{table}

\section{Proof for CDL's Conciseness}
\label{sec:sup_conciseness}

In this section, we provide a proof to demonstrate the conciseness of Conditional Declaration Language (CDL) compared with general textual descriptions.

Generally, a textual description of a geometric input can be decomposed into three components: 1) shape descriptions that depict geometric shapes, \emph{e.g.,} line segments, angles, triangles, \emph{etc.},
2) relation descriptions that reflect the positional and algebraic relationships between shape elements, and quantitative metric (\emph{e.g.}, the line segment length and the angle degree),
and 3) irrelevant words that are irrelevant to any geometric shapes or relations, \emph{e.g.,} ``a'', ``the'', \emph{etc.}

%CDL is designed with inherent constraints:\\
For a specific geometric input, let $H$, $R$, and $O$ denote the sets of all possible shapes, relations, and irrelevant words, respectively, under specific constraints or rules (\emph{e.g.,} CDL rules).
%we compare CDL to common textural descriptions from the above three aspects: 

(1) \textbf{Shape Descriptions}: 
In CDL, the shape description set $H^C$  consists of closed shapes, angles, and line segments.
There are two constraints: a) the closed shape cannot be decomposed further into other closed shapes;
b) the angle does not exist in any closed shape in the diagram; the line segment does not exist in any angle or closed shape in the diagram.
While in general textural descriptions $H^T$, there is no such constraint.
Thus, in $H^T$, in addition to elements in $H^C$, there may also exist closed shapes that can be decomposed further and angles or line segments that exist in a specific closed shape.
Consequently, $H^C \subseteq H^T$ and $|H^C|\le |H^T|$.
%Suppose a geometric diagram is composed of $|\mathcal{H}|$ different basic shapes. The set $\mathcal{H}$ consists of closed shapes which cannot be composed into other closed shapes and angles which are not in any closed shapes. 
%Note that a piece of shape description include the type of shape and the end points within this shape.

%Thus, the size of shape descriptions $H^C$ of CDL equals $|\mathcal{H}|$.
%While in common textural descriptions, any 

%CDL simply describes shapes in a diagram using a finite set of $h$ basic shapes, limiting its shape search space to $|H_{\text{C}}|=C_{h}^{1}$.
%In comparison, general text can describe any complex shape composed of these basics, resulting in a significantly larger search space:$|H_{\text{C}}|=C_{h}^{1} \le |H_{\text{T}}|\le C_{h}^{1}+...+C_{h}^{h}$.
%Thus, $|H_{\text{C}}| \le |H_{\text{T}}|$\\
(2) \textbf{Relation Descriptions}: %CDL projects relationships directly and succinctly from the diagrams or captions, ensuring CDL's search space of relation descriptions $|R_{\text{C}}|$ is non-redundant compared with $|R_{\text{T}}|$.\\
Suppose we need to depict $K$ relations to solve our problem.
For each piece of relation, there is only one possible description for such a relation in CDL. So in total, $|R^C|$ of CDL equals the number of relations $K$. 
However, there are many choices for describing the same relation in general textual descriptions. Thus, in total $|R^T|\gg K$.

(3) \textbf{Irrelevant Words}: 
As CDL is a description language with a predefined format, CDL doesn't contain any irrelevant words.
Thus, the $O^C$ for CDL and $O^T$ for general text satisfy $|O^C|=0\ll |O^T|$.

\begin{figure*}
  \centering
  \includegraphics[width=\linewidth]{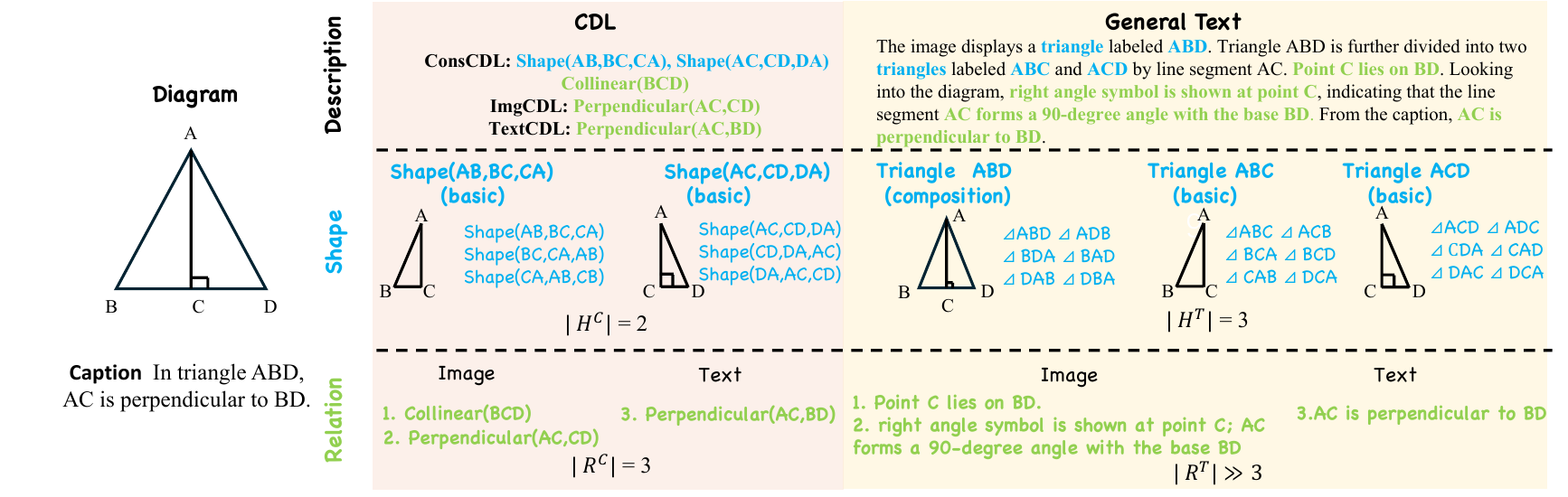}
  \caption{
  \textbf{Illustration of CDL's Conciseness.}
  A textual description of a geometric diagram can be decomposed into \textbf{\textcolor{blue}{shape descriptions ($H$)}}, \textbf{\textcolor{green}{relation descriptions ($R$)}}, and \textbf{irrelevant words ($O$)}.
  We use ``basic'' to denote the triangles that cannot be decomposed further, and ``composition'' to denote the complex triangles composed of two basics.
  Under specific constraints in the CDL rule, $|H^{\text{C}}|$, $|R^{\text{C}}|$, and $|O^{\text{C}}|$ are minimal to $|H^{\text{T}}|$, $|R^{\text{T}}|$, and $|O^{\text{T}}|$, respectively.
  Therefore, for any given geometric diagram, a CDL description is significantly more concise than a general textual description.}
\label{fig：conciseness}
\end{figure*}

Therefore, for any given geometric input, the following inequality satisfies:
\[
| \text{CDL} | = |H^C| + |R^C| + |O^T| \ll |H^T| + |R^T| + |O^T| = | \text{Text} |.
\]
This inequality proves that for a given geometric input, the number of all possible descriptions in CDL $| \text{CDL} |$ is much smaller than that in general textual descriptions $| \text{Text} |$, \emph{i.e.,} the CDL description is more concise than the general textual description.
We provide an illustration to demonstrate this in Fig.~\ref{fig：conciseness}.

\section{More Qualitative Results}
\label{sec:sup_qualitative}

In this section, we provide examples of various benchmarks, including Formalgeo-Rec-CoT, Unigeo, and Mathvista.
\begin{figure*}
  \centering
  \includegraphics[width=0.8\linewidth]{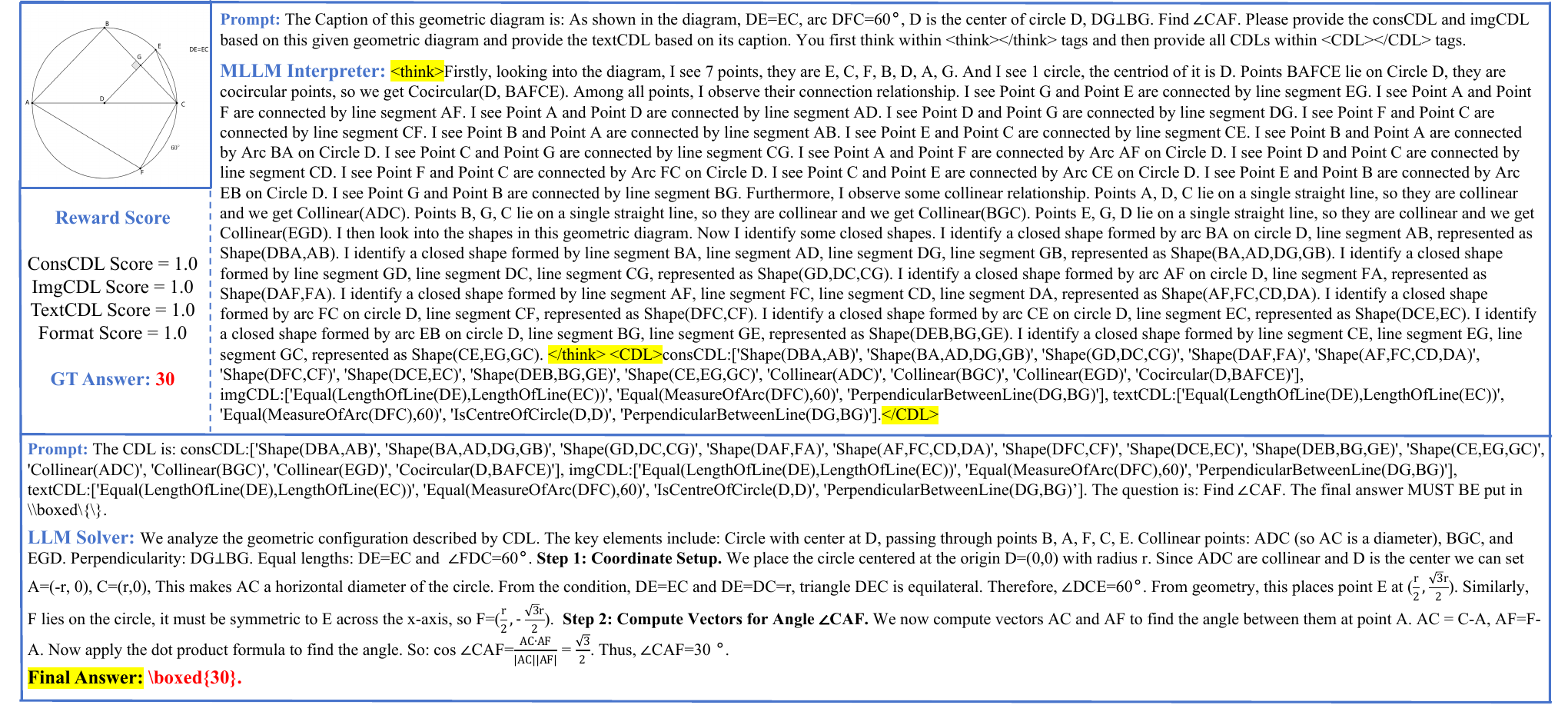}
  \caption{
  \textbf{Qualitative Result on Formalgeo7k-Rec-CoT.} }
\label{fig：visualization}
\end{figure*}
\begin{figure*}
  \centering
  \includegraphics[width=0.8\linewidth]{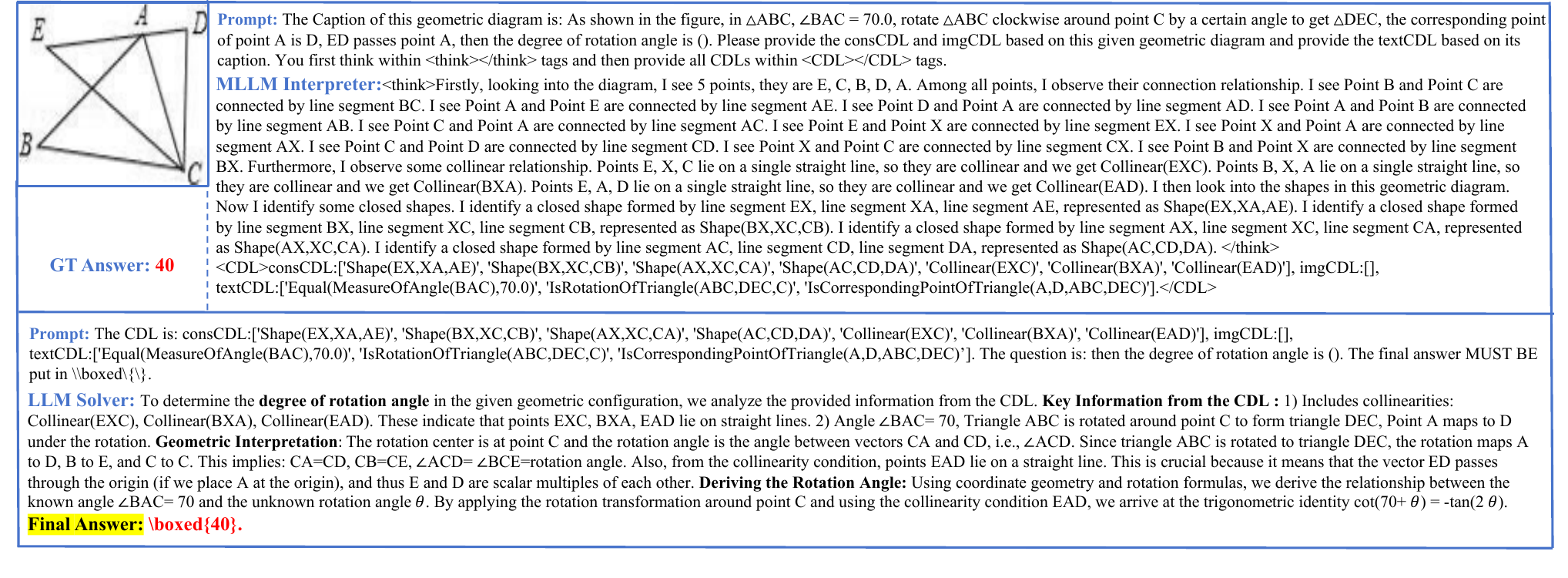}
  \caption{
  \textbf{Qualitative Result on Unigeo.} }
\label{fig：visualization}
\end{figure*}
\begin{figure*}
  \centering
  \includegraphics[width=0.8\linewidth]{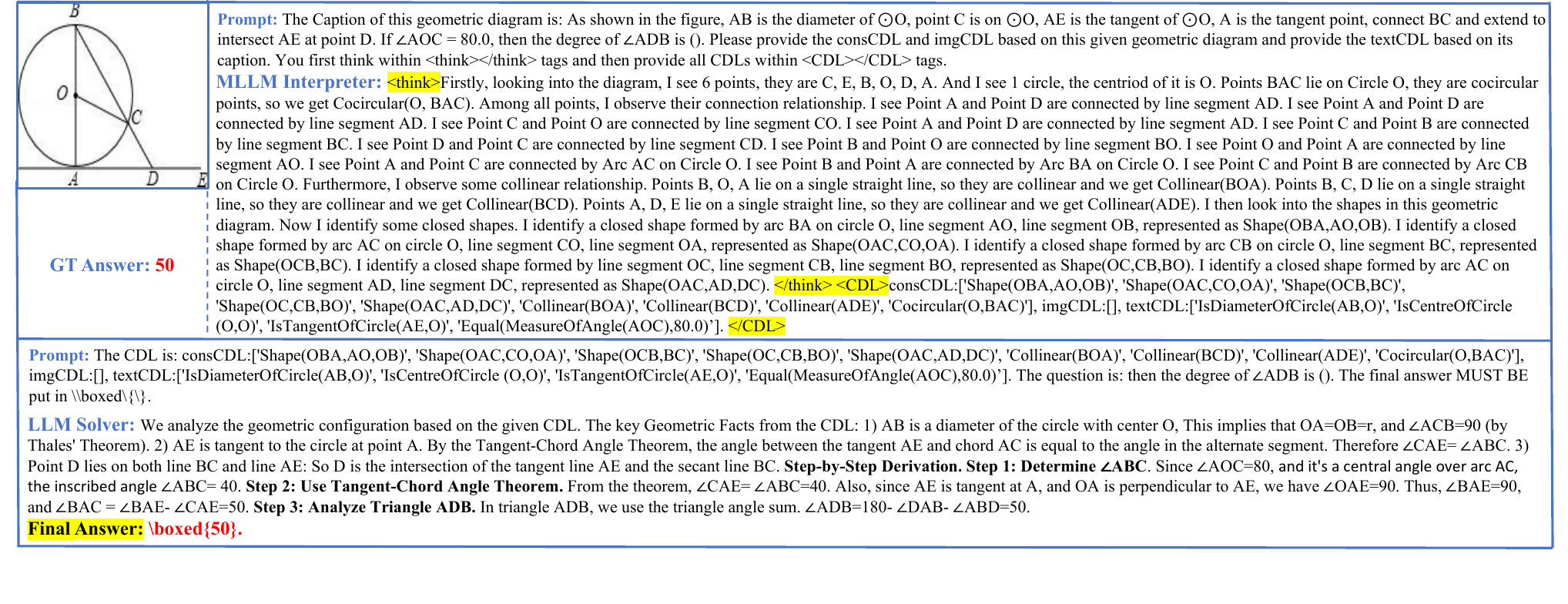}
  \caption{
  \textbf{Qualitative Result on MathVista.} }
\label{fig：visualization}
\end{figure*}

\end{document}